# On the Generalization Capability of Evolved Counter-propagation Neuro-controllers for Robot Navigation

Amiram Moshaiov and Michael Zadok

Faculty of Engineering, Tel-Aviv University, Israel
moshaiov@eng.tau.ac.il, michaelzadok@walla.co.il

**Abstract.** Evolving *Counter-Propagation Neuro-Controllers (CPNCs)*, rather than the traditional *Feed-Forward Neuro-Controllers (FFNCs)*, has recently been suggested and tested using simulated robot navigation. It has been demonstrated that both convergence rate and final performance obtained by evolving CPNCs are superior to those obtained by evolving FFNCs. In this paper the maze generalization features of both types of evolved navigation controllers are examined. For this purpose the controllers are tested in an environment that drastically differs from the one used for their training. Moreover, a comparison is carried out of results obtained by single-objective and multi-objective evolution approaches. Using a simulated case-study, the maze generalization capability of the evolved CPNCs is highlighted in both the single and multi-objective cases. In contrast, the evolved FFNCs are found to lack such capabilities in both approaches.

## 1 Introduction

In contrast to supervised learning, where generalization is commonly accounted for, generalization is rarely considered in *Evolutionary Robotics (ER)* studies [1]. This is not to say that the generalization capabilities of evolved solutions are not important in ER. As stated in [1], *the lack of methods to promote generalization of controllers curbs the application of ER to real-world robotic tasks*.

As suggested in the background section, which follows and extends the review in [1], there are only a few ER studies that concern generalization rather than just robustness. In the few available ER studies on generalization there are different generalization aspects that are considered. This paper concentrates on maze generalization in the context of robot navigation, following a benchmark problem from [2]. It concerns neuro-controllers, which are evolved to navigate a robot in a training maze environment. Following their training by evolution, the evolved controllers are examined for their navigation capabilities in a testing environment that is dramatically different from the training one.

In [2], the aforementioned benchmark problem was used to test a *Modular Neural Network (MNN)* controller approach, which is based on environment classification. According to [2], the MNN controller performed well, whereas a non-modular feed-forward approach failed to pass the test.

The MNN approach is promising from the maze generalization viewpoint due to the initial training on prototype environments. It is conceivable that each such environment results with a unique prototype behavior. Furthermore, with a proper design, the obtained set of behaviors could be typical to a general set of environments. However, the process suggested in [2] is semi-manual, as it requires designers to analyze the environment, to decide on the prototypes and to carry on the separated training of each of the modules.

At least two main questions have to be raised when considering the semi-manual approach of [2]. First, could an evolutionary method be devised to automatically create controllers, which contains prototypical information about the environment? Second, would such an automatic method have generalization capabilities similar to those demonstrated with the semi-manual MNN approach? To a large degree, these questions were positively answered in [3]. In [3] it has been suggested to automatically evolve *Counter-Propagation Neuro-Controllers (CPNCs)*. Such neuro-controllers include a self-organizing (instar) network of Kohonen [5] as a first layer and a Grossberg's outstar net [6] as the second one. In view of the special characteristics of such neuro-controllers, the evolutionary process, as proposed in [3], substantially differs from the one commonly used in ER studies. It involves not just training of a mapping from sensed information to actions. Rather, it includes organization of the sensed information into clusters based on a similarity measure, and a mapping from the obtained clusters into actions. It has been demonstrated in [3] that the proposed CPNC approach overcomes the maze generalization problem of [2] in a fully automatic manner.

While claiming to be relevant to both single and multi-objective problems, the reported generalization results in [3] have been restricted to a multi-objective evolutionary approach. This means that the reader is left to wonder if the proposed method would show generalization capabilities when the problem be formulated as a single objective one. Moreover, in [4] a revision to the CPNC evolutionary process of [3] has been suggested without testing the generalization capabilities of the revised approach.

This paper, which follows [3] and [4], aims at providing an answer to the following main question. *Does the revised pseudo-code of [4], when applied in either a single-objective or a multi-objective approach, overcome the maze generalization benchmark problem of [2]?* It should be noted that, according to [2], traditional neuro-controllers failed the aforementioned test. In addition to answering the above main question, the current work affirms that traditional feed-forward neuro-controllers fail the test in both a single-objective and a multi-objective evolutionary approach. To substantiate and to further understand the maze generalization power of the proposed CPNC approach, a comparative adaptation study is carried out. In addition, when compared with [3], this paper provides an extensive background that allows positioning this study with respect to the current state-of-the-art concerning generalization in ER, and in particular with respect to ER studies on maze generalization in navigation.

When compared with the few existing studies on maze generalization in ER, which are reviewed in the next section, it is clear that the proposed approach substantially differs from them. First, the type of neuro-controllers used here is completely different from those commonly employed in other ER studies. Second, the proposed mech-anism of generalization is based on the type of neuro-controllers that is used here. Hence, it is safe to say that a novel mechanism is proposed for maze generalization in ER studies, which may in the future be enhanced with the incorporation of other known mechanisms (e.g., co-evolving controllers and environment).

The rest of this paper is organized as follows. Section 2 provides the relevant background. It is followed by a methodology, in section 3, which outlines the experimental set-up and the evolutionary search approach. The results of the numerical simulations are presented and analyzed in section 4. Finally, the conclusions from this study are provided in section 5.

## 2    Background

### 2.1 Evolving Generalized Solutions

Robustness of evolved controllers has been an important ER research topic. One major reason is the uncertainties involved when transferring simulation results into real robots. However, as indicated in [1], robustness could be considered different from generalization. While both deal with solutions that can cope with changes from the trained situations, robustness may imply differences that are relatively small, whereas generalization could be thought as aiming at testing situations that are conceived to be substantially different from the trained one. Although *"small difference"* and *"substantial difference"* are vague terms, it appears that ER methods, which have been developed for robustness, do not necessarily promote generalization in an explicit way. With respect to the above distinction between generalization and robustness, it should be noted that some researchers may oppose this view and use the terms interchangeably.

An extensive comparison of existing ER studies and methods on both robustness and generalization can be found in [1]. Table 1 of [1], lists five ER studies that addressed generalization, however, only three of them include testing. Jakobi, in [7], considered the problem of bridging the reality gap. It is noted there that unfortunately, any real-life simulation will differ from a perfect copy of the real-world on two accounts. First, it will model only a finite set of real-world features and processes. Second, those features and processes that it does model, it will model inaccurately. Jakobi suggested conditions for successfully transferring controllers from simulators to reality, and demonstrated the proposed method. Another early method to address generalization in ER was proposed by Berlanga et al. [8]. Their method, which is termed Uniform Co-evolution, is based on co-evolving the controllers and the environment. This idea allows the evolved controllers to experience not just a particular environment; hence, generalization capabilities are expected to be achieved. Another method, which is listed in table 1 of [1], is the use of short-term-memory mechanism, in [9]. While studying various scenarios in T-mazes, [9] does not provide an assessment of the generalization capability of the method. Barate and Manzanera, [10], compared two versions of genetic programming and showed that one of them achieved generalization capabilities. A promising method to achieve generalization is to use Pareto-optimality to obtain a set of controllers with a varying trade-off among different ob-jectives. Doncieux and Mouret, [11], suggested to include behavior diversity as an objective in such a setting, and studied various initial positions of a robot. However no generalization testing is reported in [11]. In a more recent study a similar multi-objective approach has been studied with the employment of a transferability measure, to address the reality gap [12].

Clearly, table 1 of [1] misses the work in [2] on the MNN approach, where generalization was tested. More recent work that could be added to such a future table is the one in [3] on the CPNCs, which sets the base for the current study. Other ER studies that are worth mentioning are those that do not explicitly deal with the need for robustness and generalization, but discuss over-fitting, which means the opposite of generalization. For example, in [13] a modular neural-network approach is examined for evolving complex behaviors. According to [13] there is a need to ensure that each module is not over-trained. Another type of related works is that dealing with the structure of the controllers, and in-particular with the minimization of the number of neurons. With this respect, the reader is referred to the discussion on generalization in [14].

A generalization-related issue has been recently spotted concerning co-evolution. In studies on co-evolution, several methods have been utilized to measure progress during that process. According to [15], these methods measure historical progress while implying to measure global progress. In particular, the common fallacy of these methods is that they measure performance against previous opponents – that is, the sequence of successive opponents against which they have evolved. In this case, the training set is being used as a test set. Both [15], and [16] suggest methods to alleviate the aforementioned generalization-related problem in co-evolution.

Worth mentioning are also works that deal with adaptation to new environments. In such works, controllers that were trained on one or more environments are expected to gradually adapt to new environments (e.g., [17]). Although adaptation capability is not equal to generalization capability, these issues are related; the interplay between the two is yet to be studied.

The research community that deals with supervised learning uses an accepted methodology to check generalization of the learnt models. In contrast, for ER studies no accepted method to check generalization is available. Commonly, the community of researchers of ER has used the term generalization in ER studies neither with a method to measure it, nor with a clear definition of it. In [1] it has been proposed to adapt the standard three data sets methodology of supervised learning to ER applications. Such a systematic approach, to examine generalization in ER studies, appears promising but may require prohibitive computational efforts. In fact, in [1], the use of a surrogate models is suggested as a means to reduce this problem. Usually, in supervised learning each input-output pair is evaluated and data sets with known associations between inputs and outputs are available to support checking generalization. In contrast, in evolutionary learning sequences of input-output pairs are commonly evaluated. Moreover, the evaluation is achieved by interactions with the environment, which makes the search substantially harder when compared with supervised learning.

Due to the dissimilarity between supervised learning and the evolutionary learning that is typical in ER, it might be proven worthwhile to consider generalization methods from a more similar machine learning approach. A potential candidate is *Rein-*

*forcement Learning,* which has much resemblance to ER. In reinforcement learning, as in ER, sequences of steps and the associated interactions with the environment are commonly needed for the evaluation of solutions. Generalization has been a topic of interest for the reinforcement learning research community, and future ER studies on generalization might benefit from such studies. Moreover, some useful inspiration may be reached from observing reinforcement learning in Nature, as related to generalization in living creatures. Interesting to note is the recent review on the phenomenon of reinforcement learning in animals and humans, [18]. According to that review, brain over-training may cause inflexible and slow-to-adapt behaviors.

Over-training, also termed as over-fitting, is a well-known phenomenon in machine learning and in other areas that deal with extrapolation. Over-fitting is the opposite phenomenon with respect to generalization. Following the execution of their proposed ER generalization method, two important observations are made in [1] by Pinville et al. First, it has been observed that while training on a small training-set is fast, it can quickly lead to over-fitted solutions with low generalization abilities. Second, the use of a larger set is more informative but cannot be computed for each evaluated solution because of the computational cost. While not arguing with these observations, the approach here suggests that generalization could be strongly supported by organization of the sensed data into clusters, and by memorization and exploitation of the clusters by the evolved controllers.

In the current study we do not attempt to provide measures of generalization for ER, nor do we follow the procedure suggested in [1]. Rather, we have used our available computational resources for a systematic statistical examination of one case study of maze generalization, following [2]. Namely, the controllers trained in one environment are checked by observing their behavior in a testing environment that differs substantially from the training one.

When considering the different types of generalization aspects which have been accounted for in references such as [1], [2], [7] and [8], it appears that only [2] and [8] have actually focused on maze generalization, where the walls of the testing maze are dramatically changed from the training one. The use of the benchmark of [2] is most appealing here due to the similarity between the semi-manual evolutionary approach of [2] and the automated approach that is studied here. Furthermore, the benchmarking problems of [8] suffer from the use of a weighted objective, which makes the interpretation of the results complicated as compared with the success-failure characteristic of the benchmark of [2]. Consequently the current study follows that of [2].

To support understanding of the procedures used here, the following provides some background on the employed types of controllers (sub-section 2.2) and on multi-objective evolution (sub-section 2.3).

### 2.2. Feed-forward and Counter-propagation Networks

Traditionally, a large part of ER studies on robot control systems involves using *Feed-Forward Neural-networks (FFNs).* Such a controller is termed here as a *Feed-Forward Neuro-Controller (FFNC).* FFNCs are neuro-controllers that process sensory information directly into actions. Due to their feed-forward characteristics, their use in ER can be classified under behavioral robotics. They do constitute a memory. However, this memory is restricted to a non-linear mapping, which directly associates the sensory information with actions (reflexive behaviors). Hence, no explicit memory of the environment is involved when using FFNCs. In such a case, the reflections of the environment, as experienced during the evolutionary process, are implicit in the evolved mapping. As suggested in [2], and further studied here, automatic evolution of FFNC fails to produce maze generalization capability.

Recently, in [3], an alternative evolutionary neuro-control approach has been suggested. It concerns neuro-controllers that do not process sensory information directly into actions. The proposed controllers are termed *Counter-Propagation Neuro-Controllers* (CPNCs), and they are based on the well-known *Counter-Propagation Networks (CPNs)*. In contrast to FFNs, CPNs involve two mappings. First, the sensed input is mapped into a group. Namely it is associated with a cluster of input vectors. Second, each cluster is mapped into the outputs. More specifically, such a network includes a self-organizing (instar) network of Kohonen [5] as a first layer and a Grossberg's outstar net [6] as the second one. A schematic view of a CPN is given in figure 1. The scheme could be misleading as it looks as a multi-layered *Feed-Forward Network (FFN)*; yet, one should bear in mind that the neuron type and the learning rule in the first layer are different here. In the first layer each neuron involves just a weighted sum, with no self-threshold. The firing of neurons in the first layer is based on a comparison between the activation levels of those neurons. The weights of the first layer are tuned based on a similarity measure, where the sensed vector is compared with the weight vector of each of the Kohonen neurons (first layer neurons). During training, a clustering process occurs in the first layer. In contrast to the classical winner-take-all clustering procedure, where updates occur only to the weights of the winning neuron, the Kohonen procedure generally allows updates also to the weights of neighboring neurons. During testing, the first layer classifies the inputs. When serving as a controller, as done here, the weights of the Grossberg layer of the network are expected to be tuned to achieve proper mapping between the classes and the actions.

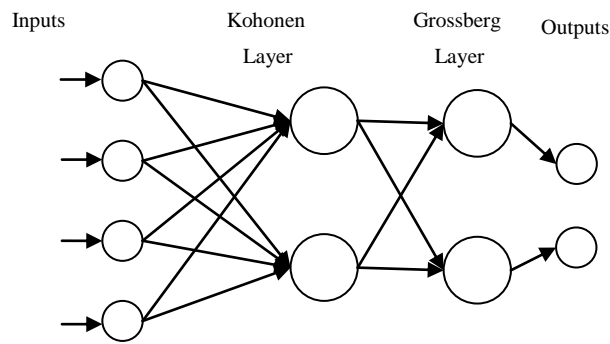

**Figure 1**: Scheme of a Counter-Propagation Network (CPN)

The original idea of mixing Kohonen and Grossberg layers, into a counter-propagation network, is attributed to Hecht-Nielsen [19]. While a promising network concept, the use of such neural-networks is not as common as that of the traditional feed-forward ones. The latter (FFNs) are simple and easy to use, whereas the former (CPNs) add the feature of classification; hence, some merit should be expected from using CPN in comparison with FFN. In [19] a comparison is provided of CPNs vs. FFNs, which are trained by the back-propagation method. It shows that there are some applications where CPNs are superior. In [20], Zupan et al. suggested that learning rates of CPNs are superior to back-propagation by orders of magnitudes. In the current study we show that CPNs are also superior to FFNs when used to evolve generalized neuro-controllers.

The generalization features of CPNs are likely to be attributed to the interpolation and extrapolation capabilities of self-organizing maps, which have been demonstrated in many machine learning studies on recognition applications. With this respect, the reader is referred to [21], which discusses the engineering applications of self-organizing maps. It is stated there that the orderliness of such input-output mapping can be utilized for many tasks including reduction of the amount of training data, speeding up learning, nonlinear interpolation and extrapolation, generalization, and effective compression of information for its transmission.

The use of self-organizing maps is not new to robotics (e.g., [22]). A more recent example is the study in [23] by Kuipers et al. on robotic learning from sensor-motor experiences. In their work they show how an agent can use self-organizing maps to identify useful sensory features in the environment, and how it can learn effective hill-climbing control laws to define distinctive states in terms of those features, and trajectory-following control laws to move from one distinctive state to another. However, their learning methodology is at the level of the individual and not at the evolution level.

Surprisingly, to the best of our knowledge, CPNs have not been used as neuro-controllers (CPNCs) in ER studies prior to [3]. Given the expected benefit of CPNCs over FFNCs, it is interesting to investigate and compare them for ER applications. As indicated in [3], the apparent advantage of using CPNCs is that, once trained, they provide generalized knowledge about the environment in the form of input classes. In regular (non-evolutionary) training of CPNs there are two phases. The first is to cluster the inputs, and the second is to create a mapping by the use of a supervised approach. Unsupervised learning, in the Kohonen self-organizing layer, is commonly based on neighborhood functions. This means that weight adjustments are done not only for the winner neuron but also to its neighboring units [5]. Due to the lack of a supervisor in ER, and due to the learning by interactions with the environment, there is a need to re-examine existing CPNs learning algorithms. In particular, there is a need to investigate various alternatives to evolutionary training of CPNs, and to compare it with other approaches. In [3], we have suggested one possible pseudo-code that can be used either with a single objective or for multi-objective evolution. The code was modified in [4], and the description of it is provided here for the sake of completeness. As noted in the introduction, the multi-objective implementation of the pseudo-code of [3], coped well with the maze generalization benchmark problem of [2]. However, the revised version of [4] was not tested for generalization, which is the focus of the current paper.

With increasing interest in cognitive robotics, the type of training in ER should shift from simple behavior-based mappings of sensors to actuators to more complex approaches. CPNs are one such possibility, which has not been investigated in the context of ER. Other neural-networks approaches, which have memory capabilities, are constantly being examined in the context of ER (e.g., [14]). Comparing such memory-based neuro-controlling approaches with the proposed CPNCs approach is left for future research.

### 2.3 Multi-objective Evolution

With the availability of *Multi-Objective Evolutionary Algorithms (MOEAs)*, e.g. [24], several ER studies employed such algorithms to obtain Pareto-optimal neuro-controllers based on contradicting objectives (e.g., [25]). Pareto-based search deals with finding Pareto-optimal set, or its numerical approximation, using dominance relation. A Pareto-optimal set includes non-dominated solutions from the feasible search space given no a-priori preferences on a finite set of objectives which are contradicting. The Pareto-front is the set of performance vectors in objective space of all solutions of the Pareto-optimal set.

The usefulness of diversity, as obtained by a Pareto-optimality approach, has been recently demonstrated, in [26] and [16], for the bootstrap problem that is common in single-objective ER. Such studies suggest that reaching diverse behaviors for one problem may produce useful (initial) solutions for another problem. The motivation to use a multi-objective approach is two-fold. First, as in [26] and [16], it provides diversity, which may help coping with numerical problems. Second, as in [25] and in similar studies, it provides useful controllers for different scenarios. In particular, similar to [25], we use the trade-off between safety and the attraction to targets to produce a diversity of controllers, with remarkable different behaviors. To obtain diverse solutions we employ NSGA-II, [24], a well-known multi-objective evolutionary algorithm, as the evolutionary search mechanism. Due to the permutation problems the use of a genetic algorithm is not recommended for the evolution of neural-networks [27]. Hence, as pointed out in [25], NSGA-II may not be the most optimal search algorithm for neuro-controllers, and a modified version, as used in [25], or more advanced algorithms, such as the MO-CMA-ES, [28], may be better. Yet NSGA-II proved to be useful for our current demonstration purposes.

The use of a multi-objective approach for generalization, has been suggested in [1], [3], and [12]. As hinted-at in the introduction, in the current study we aim to resolve, among other thing, the doubt about the source of the generalization, which was demonstrated in [3]. Namely, the generalization in [3] may be attributed to either the CPNC approach, to the multi-objective approach, or to their combination.

## 3  Methodology

To perform the current ER generalization study on CPNCs, the robot is simulated based on details from [29], [25] and [3]. The navigation problem is defined using two objective functions following [25], and is employed on both a simulated training environment and a simulated testing environment, which correspond to the benchmark problem of [2]. For the sake of completeness the details are re-described here. As in our previous studies, we concentrate on producing a simulation-based population of neuro-controllers. Such solutions may be considered as candidates for use, with adaptation, in actual testing, which is likely to be required for coping with un-modeled aspects of the simulation.

### 3.1  Simulated Robot

The robot model is based on the miniature 5.5 cm diameter Khepera robot with the following sensors. Sixteen simulated sensors are used, out of which eight simulated infra-red sensors identify obstacles (walls) and the others identify targets. The sensors are located, as pairs of an obstacle and a target sensor, at eight locations as shown in figure 2.

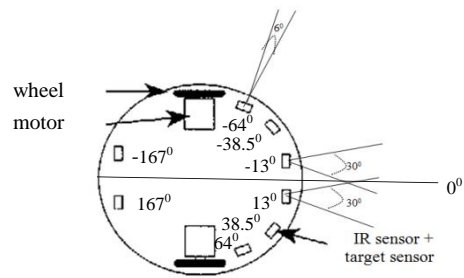

**Figure 2**: Robot and sensors

All simulated infra-red sensors have the same characteristics. The max range of any such sensor is 5cm and its span angle is $6^0$ (as shown for one of them). The target sensors have a broader span of $30^0$ (shown for the front sensors). The max range of the target sensor is simulated to be 100cm, which ensures target sensing anywhere in the maze. The sensors do not have a "blind range," and their output range is [0, 1]. The zero represents an object found at the max range, and the one is for the case when the sensed object is at the robot periphery. In the current implementation no noise is added to the simulated sensors, which is left for future research.

The simulated robot model, which involves two wheels, converts motor commands, on rotational speeds of the wheels, from the outputs of the neuro-controllers into simulated robot motions. The range of the wheel speeds is scaled into the range [-

0.5, 0.5]. The wheels radius is taken as 1cm. The time-step of the simulation is set to 5sec, and the robot moves 2.5cm per step at maximum speed.

### 3.2 Simulated Neuro-controllers

The simulated CPNCs are constructed as schematically shown in figure 1. The input layer includes the inputs from the 16 sensors. These are connected to the Kohonen layer (a hidden layer). The Kohonen layer has 9 neurons that connect to two neurons in the Grossberg (output) layer, which provides the commands to the two wheels. The reason of using a Kohonen layer of 9 neurons is that we try to compare it with the MNN approach of [2]. The 9 neurons follow the 9 classifications used in [2]. We have made an additional study, which is not reported here, to validate that 9 neurons are an optimal number for the current study. Each CPNC is defined by a real-valued vector of 163 components, out of which 162 represent connection weights. This is based on: 16 input weights multiplied by 9 neurons of the Kohonen layer + 9 connections weights from the Kohonen layer to each of the 2 Grossberg neurons. The additional component is used to determine the slope of the sigmoid for the activation functions. No bias weights are used.

To make a comparison between CPNCs and FFNCs fair, one must use a similar dimension for the search space. For this purpose the structure and the number of the neurons in the FFNCs were kept as in the CPNCs.

### 3.3 Training and Testing Environments

Both the training and testing environments, which are depicted in figures 3a and 3b respectively, are based on the environments used in [2]. According to [2], the training environment concerns nine different types of robot situations such as a wide corridor, a narrow corridor, the need to turn right/left, pass freely without walls, etc. In contrast to [2], our training approach does not use separated simple environments for a semi-manual training. Rather, we use the complex environment directly for a non-manual evolutionary training.

For the learning process, we use several targets that the robot should reach. The targets are designed to specific positions including: (90.0, 6.0), (67.5, 6.0), (60.0, 15.0), (60.0, 42.0), (90.0, 45.0), (70.5, 51.0), (60.0, 19.5), (90.0, 15.0), (75.0, 30.0), (51.0, 15.0), (45.0, 40.5), (30.0, 55.5), (3.0, 45.0). These are shown, using dots, in figure 3a. Spreading the targets aims to create an evolutionary pressure towards the different regions of the maze. In addition, we allocated a place in the maze with no targets. This supports simulating areas that are less desirable to be reached. For training, as further described in sub-section 3.5, we have used four different robot start-points located at (95, 5), (95, 45), (15, 5), (15, 45). In the two left starting points the robot is facing towards the right and vice versa.

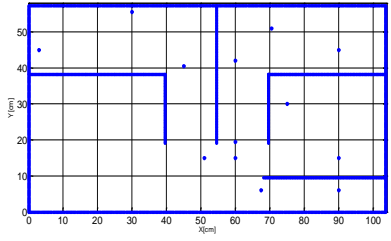 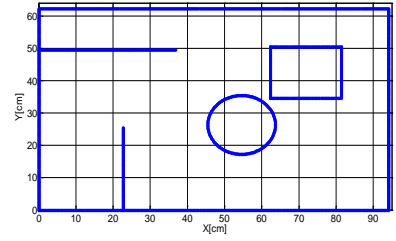

**Figure 3a:** Training environment    **Figure 3b:** Testing environment

In order to check the generality of the obtained controllers, they are tested with an unknown environment. The testing environment, which is depicted in figure 3b, is also based on [2]. For the testing case, the robot start point is at (5, 55) and it is facing right. The target point, for the testing case, is at (80, 10).

### 3.4 Objective Functions

Two fitness functions are used (marked as F1, F2). The details are similar to those used in our previous studies. The first function F1, which is based on [29], aims at fast and straight motions with obstacle avoidance without any specific destination. F1 is given as follows:

$$F_1 = \frac{\sum_{j=1}^{act\_step} V_j (1-\sqrt{\Delta v})_j (1-I)_j}{\max\_step} \quad \begin{array}{l} 0 \leq V \leq 1 \\ 0 \leq \Delta v \leq 1 \\ 0 \leq I \leq 1 \end{array}$$

Where:
- V is the absolute value of the sum of the rotational speeds of wheels. V is high when the robot is moving fast (forward or backward).
- $\Delta v$ is the absolute value of the difference between the wheel speeds.
- $1-\sqrt{\Delta v}$ is high when the robot is moving straight without making any turn during the step.
- $I$ is the normalized activation value of the sensor with the highest value. $I$ is high if the sensors perceive an obstacle.

F1 is calculated as an average over the maximum allowable number of steps of the accumulated score. The accumulation, however, is over the actual number of steps which are performed over a run of any particular controller. The function can have any value between 0 and 1, with the aim to be maximal. The second objective, F2, concerns reaching targets (e.g., food-targets). F2 is defined as follows:

$$F_2 = \sum_{j=1}^{act\_step} \frac{f_2}{\max\_step}; \quad f_2 = \begin{cases} H & \text{if hit target} \\ \frac{1}{1+d} & \text{else} \end{cases}$$

Where:
- $d$ is the distance from our robot to the nearest target among the remaining targets at the current step.
- $H$ is the score that the robot gets when it reaches a target. Here $H$ is set to 50.

Similar to F1, F2 is based on averaging of performances over the maximum allowable number of steps of the process, and summing over the actual number of steps. This reduces scores to non-moving robots at the training phase. Once a target is touched by the robot, it is eliminated (consumed). After the robot finishes touching all targets, they re-appear. Then the process of reaching targets continues as long as the robot does not reach a maximum allowable number of steps.

When used in the current environments, F1 and F2 are contradicting objectives. Here, the targets are located in narrow spaces; hence, turnings and getting close to walls are needed for reaching the target, as opposed to the case of moving in the area with no targets. The contradicting nature of the objectives implies that Pareto-optimal sets exist for the environments used here.

When the navigation problem is defined as a multi-objective problem the evolutionary learning is done using Pareto-bi-objective optimization with max F1 and max F2. In addition, two other related single-objective problems are defined, one with max F1, and one with max F2. It is noted that only the latter can be viewed as a navigation problem, since that max F2 concerns reaching targets, whereas the problem with max F1 is about obstacle avoidance without specific targeted locations. The evolutionary procedures, which are applied for all the considered problems, are described below.

### 3.5 Evolving FFNCs and CPNCs

Evolving controllers using a simulator is quite common in ER studies. The evolutionary process can be defined either by using a single-objective approach to guide the selection (e.g., [29]) or by using a multi-objective approach (e.g., [25]). As described in the above sub-section, in the current study both types of problems are considered. In particular, the well-known NSGA-II, [24], was employed when the evolutionary runs were defined as *Multi-Objective Optimization Problems (MOOPs)*. When defined as *Single-Objective Optimization Problems (SOOPs)* the runs were carried out by the same code (to maintain a fair comparison), with one difference. In the SOOP cases the performances in the other objective were kept constant. In fact, to stay impartial, all parameters of the SOOPs' runs and the MOOPs' runs, in both cases of the CPNCs and the FFNCs, were kept the same, where applicable.

Since that the focus of this study is on the evolution of CPNCs and not on the evolution of FFNCs, we do not provide here a full description for the latter. In essence, the use of the NSGA-II to evolve the FFNCs, is quite a standard ER procedure; the interested reader is referred to [30] for the details on the FFNCs' evolution procedures, which were used for the comparison study. Unfortunately, such a standard procedure cannot be used as-is for the evolutionary training of the CPNCs. A special procedure has been devised in [4], for the latter type of controllers, as represented below.

As explained in our previous studies, [3] and [4], training a CPNC requires special care due to the existence of two separated training issues. To achieve the required learning we have proposed a two-phase evolutionary search involving two interaction sequences within each phase, as depicted in figure 4, where the 1$^{st}$ phase involves the left side and vice versa.

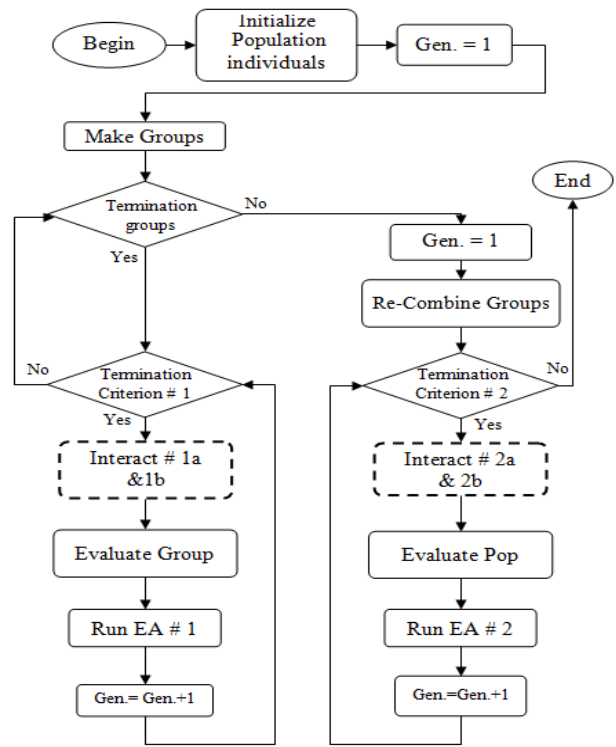

**Figure 4:** Pseudo-code Description (after [3] and [4])

The primary difference between the 1$^{st}$ phase and the 2$^{nd}$ one is that in the 1$^{st}$ phase the population is divided into groups, whereas in the 2$^{nd}$ one the groups are merged into one population. In the 1$^{st}$ phase each group has a different starting point in the environment; this separation supports learning the various classes of inputs that the entire environment contains. The proposed procedure was used for both single and multi-objective problems.

At the beginning of the 1st phase, a random population is initialized with N individuals. Each individual is a CPNC with a fixed structure, were both weights of the Kohonen layer and weights of the Grossberg layer are sought. Four groups are used in the current study, corresponding to the four start-points of the training environment (see sub-section 3.3). During the 1st phase individuals evolve only within the group. In our study we set N=56 to be divided into four groups of 14 individuals each. A *"Termination Group"* criterion is used to terminate the 1st phase of the algorithm after the completion of the evolution of the four groups. A *Termination Criterion #1* is used to terminate the evolution of each group. In the 1st phase a maximal number of generations is used (50 generations per group). At each generation each individual performs two consecutive sequences of interactions with the environment, both starting at the corresponding start-point of its group. The purpose of the 1st sequence (Interact # 1a) is to update the weights of the Kohonen layer, whereas the 2nd sequence supports updating the Grossberg layer. In the 1st sequence of interactions the updates of the Kohonen weights are done at each step of the sequence. The final Kohonen updates form *Interact # 1a* are used for the 2nd sequence, which aims to obtain the performances F1 and F2 of the individuals based on their updated version of the Kohonen layer. In the 2nd interaction sequence (*Interact #1b*) no weight update is done during the sequence of interactions with the environment. Each robot finishes the interaction (#1a and #1b) either due to an obstacle or by reaching a pre-defined number of steps (200 steps in the current implementation). Following the interactions each of the groups' individuals is evaluated using F1 and F2 based on the accumulated scoring during *Interact # 1b*.

As pointed-out above, in the current implementation the search in *EA # 1* is done, for both MOOPs and SOOPs, using NSGA-II based on [24]. Namely, following their evaluation in Interact# 1b, all the group individuals undergo a standard evolutionary cycle (*EA#1*) using NSGA-II, involving selection, crossover and mutation within the group. During the *EA # 1* evolutionary stage, the weights of the Grossberg layer are tuned as detailed further below, whereas the Kohonen layer is kept fixed (no crossover or mutation). The results from *EA #1* include offspring population (of the group) to be evaluated in the next group generation of the 1st phase. The entire 1st phase is repeated for each group until all groups are evolved for the pre-defined number of generations as described above under termination criterion # 1. Presumably, at the end of the 1st phase the genetics of each group include some capability to move in the environment based on classes learnt in the Kohonen layer. However, each group is expected to be biased by its starting point. Therefore, the 2nd phase (right column of figure 4) is devised to merge and improve the genetics from all the groups.

Once the 1st phase is terminated, the 2nd one starts with a new counting of the generation number, and with a new termination criterion. The 2nd phase starts with uniting the groups into one population. During *Interact # 2a and # 2b* updates are done similar to their counterparts of the 1st phase. Following *Interact # 2b*, F1 and F2 are calculated for each individual based on the accumulated scoring from the interaction steps. When the individuals of the entire population of the current generation have been evaluated the population undergoes an evolutionary cycle, in *EA#2*, of selection, crossover and mutation. Similarly to EA#1, in the current implementation *EA#2* is based on NSGA-II. However, it is noted that in *EA # 2* we employ a special mating procedure as described further below. A *"termination criterion # 2"* is used to terminate the $2^{nd}$ phase of the evolution. In our study a maximal number of generations is used (currently 250). Namely, when adding the generations from the $1^{st}$ phase, a total of 300 generations is employed for the entire evolution.

The above description summarizes the entire evolutionary procedure that has been devised for the evolutionary training of CPNCs. The use of *interact # 1a and #2a* provide the necessary unsupervised learning updates of the Kohonen weights during the life-time of the individual. These step-based updates, at the current implementation, involve the well-known winner-take-all procedure, using the Euclidean distance between the instantaneous input vector and the Kohonen weight vectors and an exponentially decaying learning rate [30]. No neighboring Kohonen neurons are updated in the current implementation.

The above description will be incomplete without some clarifications concerning the crossover operations that were implemented. As noted above, during the $1^{st}$ phase, the weights of the Grossberg layer are tuned, whereas the Kohonen layer is kept fixed (no crossover or mutation). At that phase, for the recombination in the Grossberg layer we used 100% probability. In the upper part of figure 5 two parent neurocontrollers are schematically shown, and their corresponding offspring are shown in the lower part of the figure. As typically depicted in figure 5, we employed: (Wyd',Wya')=SBX(Wyd,Wya); (Wzd',Wza') = SBX(Wzd,Wza). The mutation in the Grossberg layer is done with polynomial mutation. The interested reader is referred to [31] for details on these operations.

The need for a special mating procedure at the $2^{nd}$ phase is due to the danger of mating between Grossberg weights that are associated with different classes. Such a mating is likely to ruin the mapping between classes to actions that were learnt during the $1^{st}$ phase. Hence, in *EA # 2* crossover is done by comparing classification neurons (Kohonen neuron weight vectors) in the mating parents to ensure that crossover is done among weights that are connected to Kohonen neurons that represent the "same" class. First, one neuron of one of the parents is selected, and crossover is performed with the closest neuron in the second parent. Next, another neuron of the first parent is selected and crossover is performed with the closest neuron that wasn't selected before. This procedure is continued for the rest of the neurons of the Kohonen layer. In the above, the term crossover between two neurons means that the Grossberg weights from the neurons are crossed-over using the SBX approach. The crossover operation is used in the $2^{nd}$ phase with a chance of 50%, as opposed to the 100% used in the $1^{st}$ phase (based on trial and error). No mutation is used in the Kohonen layer. The mutation in the Grossberg layer is done with polynomial mutation.

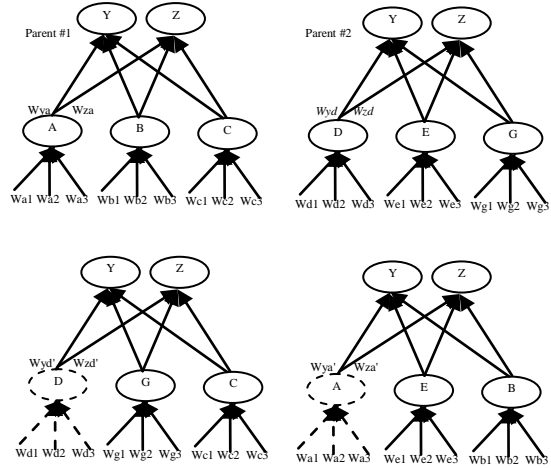

**Figure 5:** Mating and mutating CPNCs

Referring again to figure 5, assume that among neurons D, E and G, of the right parent, neuron D has the closest weight vector (Kohonen layer) to the weight vector of neuron A of the left parent. Similarly assumes G is the closest to B and E is the closest to C. Crossover occurred only for the first two cases due to the 50% chance for crossover. It is important to note that, in contrast to the $1^{st}$ phase, the crossover operation of the $2^{nd}$ phase does not keep the Kohonen layer fixed. Overall, the proposed procedure changes the statistical occurrences of the learnt classes (Kohonen weights) from generation to generation, while also tuning the mapping to actions (Grossberg weights).

## 4  Experimental Study

### 4.1 Preliminary Notes

The experimental setup has been described in the previous section. It includes: the simulated robot, sensors and actuators (described in sub-section 3.1), the neuro-controllers (described in sub-section 3.2), the training and testing environments (described in sub-section 3.3), the objective functions (described in sub-section 3.4), and the evolutionary procedures (described in subsection 3.5).

This section contains the experimental results and their comparisons and analyses. It includes comparisons not just among the obtained results, but also a qualitative comparison with the published results of [2] (as applicable).

Sub-sections 4.2 and 4.3 provide comparisons between the CPNC and FFNC methods in the training and testing environments. Overall, six cases are studied including: two different single-objective uses of CPNCs, and their repetition with the use of FFNCs, as well as the multi-objective uses of CPNCs and FFNCs. As indicated in the introduction, this paper aims at providing an answer to the following main question. *Does the revised pseudo-code of [4], when applied in either a single-objective or a multi-objective approach, overcome the maze generalization benchmark problem of [2]?* With this respect, the training results that are provided in sub-section 4.2 serve only as complementary information. The answer to the main question is primarily based on the testing results provided in sub-section 4.3. The findings in sub-section 4.3 are followed by sub-section 4.4, which provides some further information, on the maze generalization capabilities of the CPNCs, by way of an adaptation study.

Each executed run involved simulations with 200 steps, and lasted for 50+250 (300) generations. The statistics are based on 30 runs per each test. Other parameters used for the runs are detailed in sub-section 3.5. Due to the low number of re-running the experiments, the medians are preferred over the averages when comparing the graphs. In the following sub-sections the statistical results are given in a "boxplot" form (also known as a box-and-whisker plot). In the plots, the bottom and top of the boxes are the $25^{th}$ and $75^{th}$ percentiles (the lower and upper quartiles, respectively). The lines and the stars in the boxes are, respectively, the median and the average values. The whiskers represent the lowest datum still within 1.5 IQR (Inter-Quartile Range) of the lower quartile, and the highest datum still within 1.5 IQR of the upper quartile. Any results not included between the whiskers, are shown as outliers by dots. The horizontal axis in the performance graphs is the generation number which is given in jumps of 10 generations between each subsequent column. The results for CPNCs are presented only from the sixth column. This is due to the initial focus, during the first 50 generations, on classification rather than on control.

### 4.2 CPNCs and FFNCs in the Training Environment

This section concerns the performance of evolved neuro-controllers in the trained environment. As such, it serves just to provide reference data, and it does not attempt to provide an answer on the maze generalization capabilities of the controllers. The comparison given here is between the CPNC and the FFNC methods for both single and multi-objective problems, namely for SOOPs and MOOPs. The resulting SOOP performances, in the training environment, are depicted in figures 6 to 8. Each figure includes a comparison between CPNCs and FFNCs with a specific objective. Figures 6 and 7 present comparisons with F1 and F2, respectively. It can be easily observed that the convergence with CPNCs is faster than with FFNCs. Moreover, after 300 generations the median of the CPNCs results is either practically equal (case of F1) or higher (case of F2) than the median of the FFNCs results. In the F1 objective case, the location and the sizes of the boxes (e.g. 25%-75% group of results) are better (higher and smaller, respectively) in the CPNC method. Yet, the entire spread (between 0%-100%) of the results is worse (larger spread in the CPNC method). In the F2 objective, the sizes of the boxes are larger in the CPNC method but the medians are better than the FFNC method. Here also, the results converge better but the spread is worse.

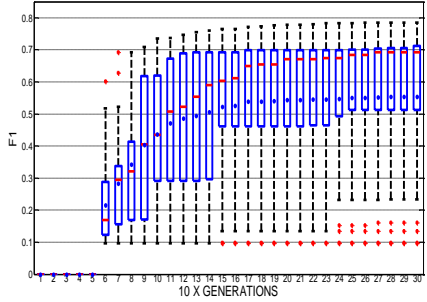
**Figure 6a:** CPNCs - SOOP - F1

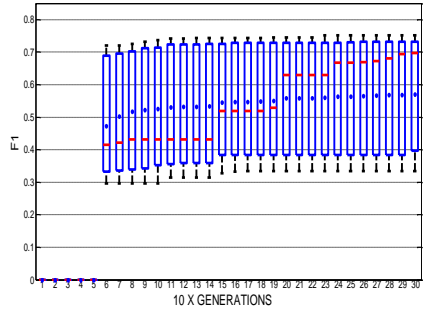
**Figure 6b:** FFNCs - SOOP - F1

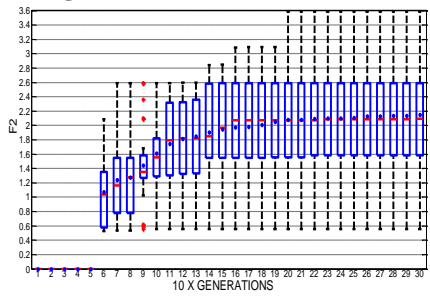
**Figure 7a:** CPNCs - SOOP - F2

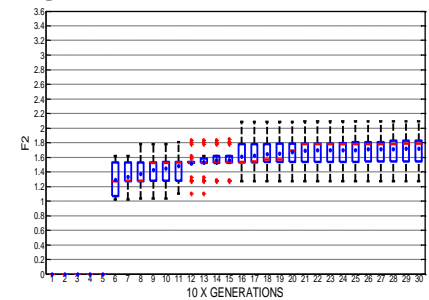
**Figure 7b:** FFNCs - SOOP - F2

Figures 9 and 10 show the paths for the controllers with the best F1 and the best F2, respectively. When observing figures 9a and 9b (Best F1 controller), recall that the concept behind this controller is to have a safe-straight moving robot with maximal speed, on the expense of reduced target collection abilities. This means that the robot is expected to be attracted to spacey areas, where it can move straight and far from obstacles, rather than to the targets. The shown paths, for both CPNC and FFNC, clearly avoid narrow areas where most of the targets are. The best F1 controllers "prefer" using the available steps on the larger, hence safer, yet empty room. Comparing the paths of figure 10a and 10b with those of figures 9a and 9b, it becomes evident that the best F2 controllers are much less safe as they run the robot into narrow places, while striving to reach all targets. As shown further below, the nature of these controllers is observable also in the testing environment.

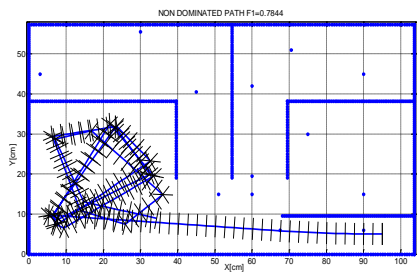
**Figure 9a:** Path of CPNC - SOOP - F1

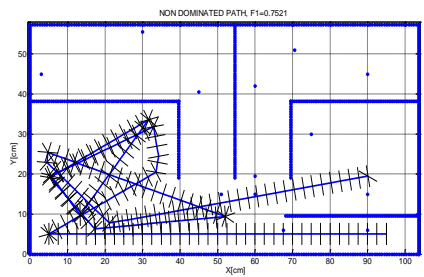
**Figure 9b:** Path of FFNC - SOOP - F1

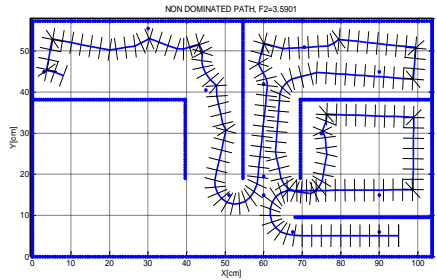
**Figure 10a:** Path of CPNC - SOOP - F2

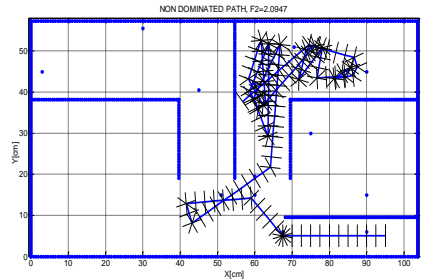
**Figure 10b:** Path of FFNC - SOOP - F2

Figures 12a and 12b present the results when the runs are done as MOOPs; the performance vectors from thirty Pareto-fronts are shown for the CPNCs and for the FFNCs, respectively. The horizontal and vertical axes are F1 and F2 respectively (max-max optimization). Each front, which was obtained by one run, is depicted by using the same shape and color for all performance vectors of the front. However, given the expected difficulties in identifying the individual fronts in the figures, especially in a non-digital black and white version, we note that none of the individual fronts is fully equal to any of the other fronts. Moreover, none of them is equal to the combined front from all the runs. In order to compare the fronts quantitatively, we calculated, for each front, the area between the front and main axes (S-measure).

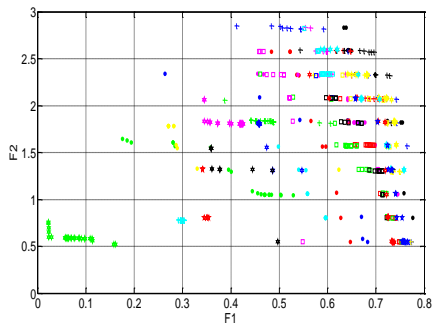
**Figure 12a:** CPNCs - Pareto fronts

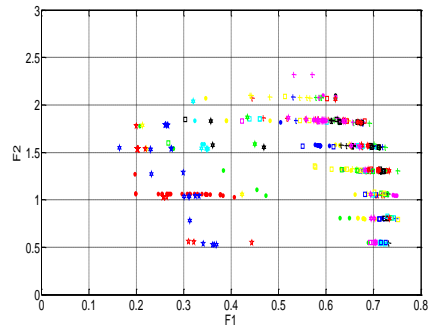
**Figure 12b:** FFNCs - Pareto fronts

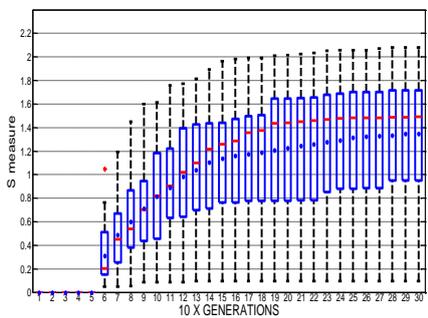
**Figure 13a:** CPNCs - S measure

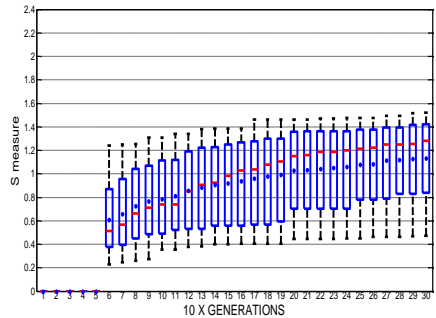
**Figure 13b:** FFNCs - S measure

As clearly observed from figures 12 and 13, there is a strong scatter in the results. Namely, the repeatability of the training results is poor. This phenomenon is due to the definition and behavior of the objective functions, and not due to the training scheme. It appears as a local Pareto problem. Due to the large dimension of the search space it is hard to explore it in-depth. Due to the lack of an analytical solution, a large number of runs would be required to increase the confidence on reaching the global front. Alternatively, NSGA-II should be replaced with a better search mechanism. Yet, due to the lack of an analytical solution, the best front obtained is always just a guess. For the purpose of this comparative study we are interested in "satisficing" solutions, which are obtained under the same search approach. Hence this phenomenon is acceptable here. In fact, the best front that is accumulated, using dominance relation among all 30 fronts of the different runs, has been sufficient for the purpose of this study. As described further below, the obtained controllers, which are associated with the accumulated front, are satisfactory for the testing environment. Namely, they provide a satisfying answer to the main maze generalization question that is posed here.

From figures 13a and 13b it is clear that the convergence in the case of the CPNCs is faster than in the case of the FFNCs. As seen in figure 13a more than 90% of the maximal value of the median was reached after 170 generations by the CPNC approach, whereas from figure 13b it is clear that more than 300 generations are needed by the FFNC approach to reach that median value. Moreover, even when considering the top part of the boxes in figure 13b, one may easily observe that the value reached at 300 generation by the FFNC approach is less than the median value reached by the CPNC approach at 190 generations.

The location and the sizes of the boxes (e.g. 25%-75% group of results) are better (higher and/or smaller, respectively) in the CPNC method although the entire spread (between 0%-100%) of the results is worse (larger spread in CPNC method). It should be noted that the median of the CPNC results (1.55) is significantly higher than the median of the FFNC results (1.3).

The primary numerical results of both MOOP & SOOP experiments are summarized in table 1. In all the experiments, the median values of the CPNC results were higher or equal to the FFNC approach and the boxes (25%-75% of the results) were higher. It should be noted that these results were obtained, due to computational resources, for runs with 300 generations. Presumably, if more generations had been allocated, then the numbers would change and perhaps the final results would reach similar values. Nevertheless, the convergence rate of the CPNCs would still be better as compared with the FFNCs.

|  | **CPNCs SOOPs** | **FFNCs SOOPs** | **CPNCs MOOPs** | **FFNCs MOOPs** |
|---|---|---|---|---|
| **F1– median** | 0.7 | 0.7 | ---- | ---- |
| **F1– [25%-75%]** | [0.5,0.7] | [0.4,0.7] | ---- | ---- |
| **F2– median** | 2.1 | 1.8 | ---- | ---- |
| **F2 –[25%-75%]** | [1.6,2.6] | [1.55,1.85] | ---- | ---- |
| **S – median** | --- | --- | 1.55 | 1.3 |
| **S – [25%-5%]** | --- | --- | [0.95,1.7] | [0.85,1.4] |

**Table 1:** Summary of Results

Overall, the CPNC approach seems to be better than the FFNC approach in the training environment. Yet, due to the large variances and the small number of samplings the above observations require further analysis. To highlight the statistical significance of our results we have conducted a Wilcoxon test on the statistical values of F1 and F2 as obtained at 300 generations by the two approaches. For the F1 case neither FFNC nor CPNC can be declared superior, which corresponds to the observed equal values of their medians. For the case of F2 the CPNC can be declared superior with the p-value of 0.013.

### 4.3 CPNCs and FFNCs in the Testing Environment

As described in section 3. 3 the testing problem differs from the training one not just by the structure and shape of the walls, but also by the target setting. Here we aim at reaching only one target (right-bottom) from a far location (left-top). Figures 14a and 14b show the obtained paths of the best F1 and the best F2 CPNCs (respectively) as obtained in the MOOPs training cases. Figure 15 shows the path that was created by the CPNC, which was obtained by training CPNCs using SOOP with the F2 objective. As seen for the three cases shown in figures 14a, 14b and 15, these tests start at the upper left point, inside the narrow corridor. In all these cases, the trained controllers coped well with the testing challenge; reaching the far location of the goal is achieved in all three cases. It is important to note that this navigation achievement was reached by the trained controllers without any further adaptation to the testing environment. The safest controller of figure 14a shows "less confidence" inside the corridor. It attempts to avoid walls (behave as safe as possible), while trying to be fast. Apparently, this results in an unstable behavior. The less safe and target-oriented controller, of figure 14b, shows a more stable behavior inside the corridor. The target-oriented controller of figure 15 shows average behavior. It appears to behave better than the safest controller but not as good as the controller of figure 14b, which was obtained by the Pareto approach. When reaching the end of the corridor both the MOOP controller, of figure 14a, and the SOOP controller, of figure 15, avoid entering the narrow space between the rectangle and the side wall. They turn into the relatively spacey zone but are still drawn towards the target, which is located at the lower-right-side of the environment. The controllers eventually reach the target. The target-oriented controller of figure 14b reached the target in a much shorter, but in a less safe way, as compared with the controllers of figures 14a and 15.

It should be noted that the navigation success shown in figure 14a, for the case of F1, might raise a question. This is due to the fact that F1 in itself does not provide an incentive for the evolved controllers to reach the goal. Yet, one should bear in mind that the F1 controller was obtained by defining a MOOP with both objectives.

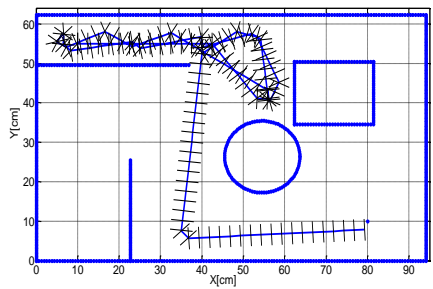 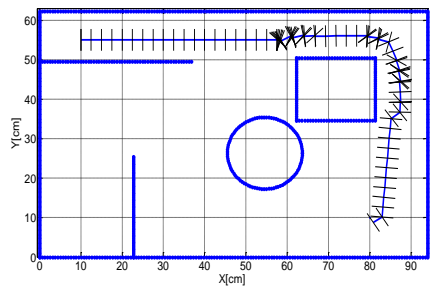

**Figure 14a:** Path of CPNC MOOP - best F1    **Figure 14b:** Path of CPNC MOOP - best F2

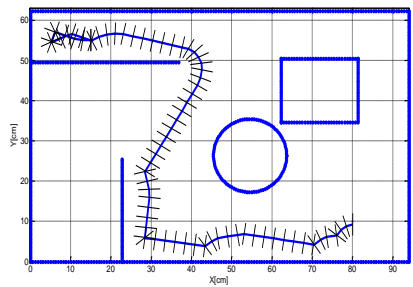 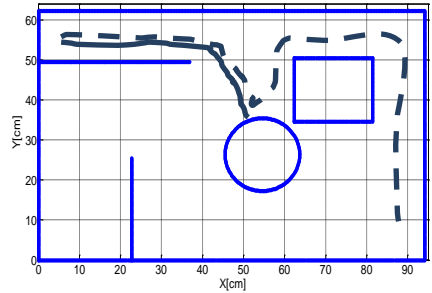

**Figure 15:** Path of CPNC SOOP - F2    **Figure 16:** MNN vs. FFNC (after [2])
Dashed/Solid line: MNN/ FFNC

The paths of the CPNCs and FFNCs, which were obtained using SOOP with the F1 objective, are not shown. They have not reached the target. This is not surprising since that these controllers were trained with no incentive to reach any target. However, while trained to reach targets, by either a SOOP-F2 or by a MOOP, *none of the FFNCs reached the target in the test-case!*

For a qualitative comparison, figure 16 re-illustrate the results that were reported in [2] on both the use of MNN and a single feed-forward network, As seen in figure 16, the MNN controller was able to reach the target, after switching its behavior close to the rectangular and circular obstacles. In contrast, the regular controller failed to get to the target, and got stuck close to those obstacles. Clearly, the MNN controller in figure 16 has a similar behavior to that achieved by the CPNC of figure 14b. Both passed the narrow corridor between the rectangle and external wall, towards the target. However, it appears that the CPNC has a better performance.

When considering the results, which are described in this sub-section, one may clearly conclude that the main question raised in this study can be positively answered. Namely, the CPNCs controllers, which were trained using the evolutionary procedure suggested in [4] and detailed here in sub-section 3.5, pass the maze generalization benchmarking test both when trained in a MOOP and in a SOOP case. The results also affirm that traditional feed-forward neuro-controllers failed the test in both a single-objective and a multi-objective evolutionary approach. It is important to highlight the statistical data that confirm the above conclusions. First it should be noted that the reported success has been confirmed not just with a few controllers, but with a substantial number of CPNCs from the different runs. Similarly, the conclusion regarding the failure of the FFNC approach has been verified with all solutions from the combined front from all the 30 runs.

To further analyze the results, one may suggest that the success of the CPNC approach, as compared with the failure of the FFNC approach, is correlated with the significantly better navigation performance of the formers as revealed in the F2 results of subsection 4.2. Yet, it was also shown that FFNCs perform equally good with respect to F1, as compared with CPNCs, which does not correlate well with the fact that the CPNCs have shown generalization capabilities even with the best F1, while the FFNCs have not.

As shown here, the successful CPNCs did not require any adaptation to the testing environment. To get a better insight on the maze generalization capability of the trained CPNCs, an adaptation study is carried out as detailed in the following subsection.

**4.4 Adaptation to Testing Environment**

In this section, adaptation of the trained CPNCs to the testing environment is studied and compared with the evolution in that environment of un-trained random population of CPNCs. Here the accumulated Pareto-optimal set of the $1^{st}$ environment is used as an initial population to be further optimized and adapt to the second environment. This population is termed pre-trained and the second, which has no pre-training, is termed random. Both populations include 56 individuals. Figures 17 (a, b) and figures 18 (a, b) provide results for the SOOP and MOOP cases, respectively. Figures 17a and 18a show the F1 statistics for pre-trained initial populations, whereas the statistical results for the random cases are shown in 17b and 17c. Figure 19a (SOOP) and figures 19b (MOOP) show the fraction of controllers that reached the target. The figure contains two curves: The first, marked by the green (dashed) curve, is of the random controllers. The second, marked by the blue (continues) curve, is of the pre-trained ones.

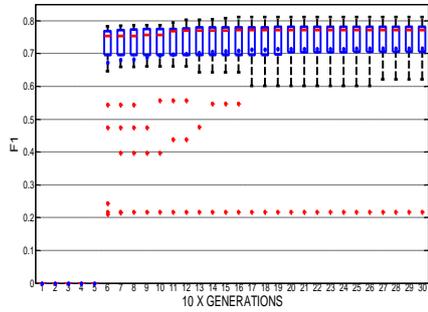
**Figure 17a:** CPNC SOOP F1: Pre-trained

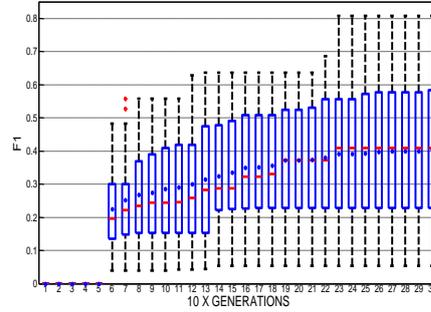
**Figure 17b:** CPNC SOOP F1: Random

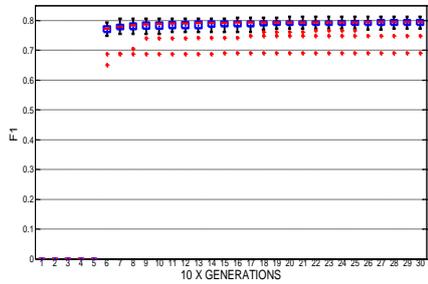
**Figure 18a:** CPNC MOOP F1: Pre-trained

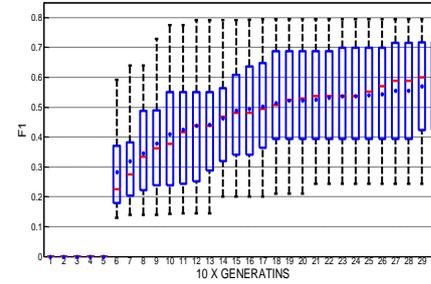
**Figure 18b:** CPNC MOOP F1: Random

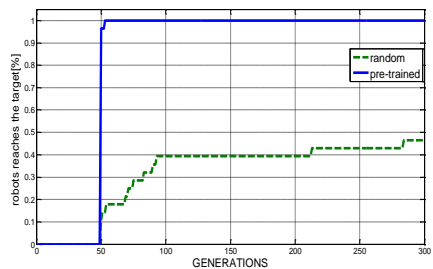
**Figure 19a:** CPNC SOOP F2: Pre-trained vs. Random

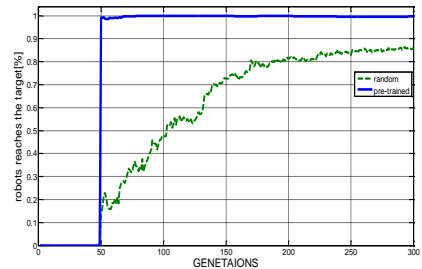
**Figure 19b:** CPNC MOOP F2: Pre-trained vs. Random

As expected, the pre-trained controllers in both SOOP and MOOP are much superior as compared with the random ones. They are faster, have full successes of reaching the target, and have better convergence. In fact, they appear to be optimal almost from the start, which is counted after the first phase of the training at 50 generations (the graphs present the 300 generations as summed from the two phases (50+250)). A comparison between the SOOP and the MOOP for the random case reveals that in this demonstration, the MOOP approach helps to get better results (higher medians in F1 results, higher percent of controllers reaching the target in the F2 results).

One may argue that there is a major flaw in the above comparison. On the one hand, the pre-trained solutions have already had 300 generations to evolve and are now given additional 300 generations to further adapt (600 generations in total). On the other hand, the randomly started ones have only 300 generations to adapt to this testing environment. Hence, the fairness of the comparison is arguable. With this respect the reader is referred to a similar discussion concerning transfer learning in [32]. According to [32] one may consider the spent cost as "sunk cost," in particular when the goal is to effectively re-use past knowledge. The case here could be viewed as such. It is in essence a case of knowledge transfer by way of the use of genetics. The training case is a source task and the testing case is a target task. The target task is novel and past knowledge is re-used. When viewed in these terms the comparison could be argued as fair. In fact such a comparison is commonly done in a large part of research in transfer learning [32]. Moreover, even without the assumption of "sunk cost," when considering the results in figures 17-19 it can easily be observed that after 300 generations of training the random solutions, the obtained values are much lower when compared with the values obtained by the pre-trained initial population at the start of the shown performance (after 50 generations of the $1^{st}$ phase). The latter comparison, which accounts for the pre-training cost, cannot be argued as unfair since that the resources invested in both populations are fairly similar, while the performances are fairly different.

The above findings suggest that training solutions from random on the testing environment require substantial computational efforts, whereas such efforts can be saved by the use of the pre-trained CPNCs. The pre-trained ones show excellent adaptation capabilities, which is not surprising given their maze generalization capabilities.

## 5      Conclusions

Regardless of its significance, maze generalization capabilities of evolved navigation controllers have hardly been studied by ER researchers. This paper shows that the evolutionary process suggested in [4] provides Counter-Propagation Neuro-Controllers (CPNCs) that cope well with the maze generalization benchmark of [2]. The aforementioned success is achieved in both single and multi-objective settings. These findings eliminate the doubt regarding the source of the generalization, which was originally demonstrated in [3]. Namely, the generalization in [3] can be attributed to either the CPNC approach, to the multi-objective approach, or to their combination. From the current results one can state that since that generalization was observed not just with the multi-objective case but also with the single-objective one, then the sole reason for generalization must be the use of the CPNC approach.

The clustering that occurs in the first layers of the CPNCs is the unique element of the proposed approach. The current results, which also include the failure of the FFNCs to cope with the problem, suggest that the clustering is the main reason for the maze generalization capability of the CPNCs. Inherent to clustering is the creation of prototypes of the sensed information, which have the ability to extrapolate and interpolate. As in [2], it was also found here that traditional feed-forward neuro-controllers failed in the testing case, whereas similar to the modular method of [2], the proposed CPNC approach was able to produce successful maze generalization. Moreover, as already mentioned here, we have made an additional study, which is currently not reported, to validate that 9 neurons are an optimal number for the current CPNC study. It is noted that similar conclusion has been reached in [2]. In fact, the ingredients of both the modular and the CPNC approaches are the same; hence the similarity of the results is not surprising. The difference is primarily that here the approach aims at a fully automated procedure. The use of CPNCs with a fixed structure, as done here, is not sufficient with this respect. For the method to be considered as suitable for maze generalization, the simultaneous evolution of the CPNCs structures should also be investigated. Moreover, in the current implementation of the CPNCs it has been observed that there is no need to adjust the weights of neighboring neurons, hence only the weights of the winner are adjusted. The further development of the proposed approach would certainly benefit from the use of a Kohonen layer that includes neighborhood considerations among similar neurons.

Future studies should also include: (a) exploitation of the proposed scheme for more challenging problems, (b) studying the classification scheme as related to the complexity of the environment and problem, and (c) actual implementation in physical environments. As already pointed-out in the background section, other neural-networks approaches, which have memory capabilities, are constantly being examined in the context of ER. Comparing such memory-based neuro-controlling approaches with the proposed CPNCs approach is also left for future research.

As a final note, we remind the reader that in the background of this paper the relation between generalization in ER and generalization in reinforcement learning is pointed-out. It appears important for the ER research community to examine such studies on generalization, which are carried out by both the machine learning research community and by researchers in behavioral psychology. As a part of the ER research on generalization, the future development of the CPNC approach would probably benefit from such studies, and from hybridizations with other generalization approaches such as those of [1], and [8]. Finally, it should also be pointed out that formalizing tasks relations and similarities, should be at the front of future ER studies on generalization. With this respect, future ER generalization studies could benefit from advancements in the field of transfer learning (e.g., [32]).

**Acknowledgement**

The authors would like to thank the anonymous reviewers for their most valuable remarks, which contributed to the writing of this paper.